\documentclass[manuscript,screen]{acmart}
\usepackage{makecell}
\usepackage{array}
\usepackage{booktabs}
\usepackage{multirow}
\usepackage{tikz}

\AtBeginDocument{%
  }

\setcopyright{acmlicensed}
\copyrightyear{2018}
\acmYear{2018}
\acmDOI{XXXXXXX.XXXXXXX}

\acmConference[Conference acronym 'XX]{Make sure to enter the correct
  conference title from your rights confirmation emai}{June 03--05,
  2018}{Woodstock, NY}
\acmISBN{978-1-4503-XXXX-X/18/06}

\begin{document}

\title{Enhancing Transferability of Adversarial Attacks with GE-AdvGAN+: A Comprehensive Framework for Gradient Editing}


\author{Zhibo Jin}
\authornote{Both authors contributed equally to this research.}
\email{zhibo.jin@student.uts.edu.au}
\orcid{0009-0003-0218-1941}
\affiliation{%
  \institution{The University of Technology Sydney}
  \city{Sydney}
  \state{NSW}
  \country{Australia}
}
\author{Jiayu Zhang}
\authornotemark[1]
\email{zjy@szyierqi.com}
\orcid{0009-0008-6636-8656}
\affiliation{%
  \institution{Suzhou Yierqi}
  \city{Suzhou}
  \state{Jiangsu}
  \country{China}
}

\author{Zhiyu Zhu}
\authornotemark[1]
\email{zhiyu.zhu@student.uts.edu.au}
\orcid{0009-0009-0231-4410}
\affiliation{%
  \institution{The University of Technology Sydney}
  \city{Sydney}
  \state{NSW}
  \country{Australia}
}

\author{Chenyu Zhang}
\email{zhangchaya6@gmail.com}
\affiliation{%
}

\author{Jiahao Huang}
\affiliation{%
 \institution{The University of Sydney}
 \city{Sydney}
 \state{NSW}
 \country{Australia}}

 \author{Jianlong Zhou}
 \email{jianlong.zhou@uts.edu.au}
\affiliation{%
  \institution{The University of Technology Sydney}
  \city{Sydney}
  \state{NSW}
  \country{Australia}
}

 \author{Fang Chen}
 \email{fang.chen@uts.edu.au}
\affiliation{%
  \institution{The University of Technology Sydney}
  \city{Sydney}
  \state{NSW}
  \country{Australia}
}

\renewcommand{\shortauthors}{Jin et al.}

\begin{abstract}
Transferable adversarial attacks pose significant threats to deep neural networks, particularly in black-box scenarios where internal model information is inaccessible. Studying adversarial attack methods helps advance the performance of defense mechanisms and explore model vulnerabilities. These methods can uncover and exploit weaknesses in models, promoting the development of more robust architectures. However, current methods for transferable attacks often come with substantial computational costs, limiting their deployment and application, especially in edge computing scenarios. Adversarial generative models, such as Generative Adversarial Networks (GANs), are characterized by their ability to generate samples without the need for retraining after an initial training phase. GE-AdvGAN, a recent method for transferable adversarial attacks, is based on this principle. In this paper, we propose a novel general framework for gradient editing-based transferable attacks, named GE-AdvGAN+, which integrates nearly all mainstream attack methods to enhance transferability while significantly reducing computational resource consumption. Our experiments demonstrate the compatibility and effectiveness of our framework. Compared to the baseline AdvGAN, our best-performing method, GE-AdvGAN++, achieves an average ASR improvement of 47.8. Additionally, it surpasses the latest competing algorithm, GE-AdvGAN, with an average ASR increase of 5.9. The framework also exhibits enhanced computational efficiency, achieving 2217.7 FPS, outperforming traditional methods such as BIM and MI-FGSM. The implementation code for our GE-AdvGAN+ framework is available at \hyperlink{https://github.com/GEAdvGANPP}{https://github.com/GEAdvGANPP}.
\end{abstract}



\received{20 February 2007}
\received[revised]{12 March 2009}
\received[accepted]{5 June 2009}

\maketitle

\section{Introduction}\label{introduction}
The concept of adversarial attacks first emerged in the field of machine learning and has gained increasing attention with the rapid development of deep learning technologies and their widespread application across various domains. In particular, deep learning models have demonstrated remarkable capabilities in areas such as image recognition~\cite{li2022research,chen2019looks,jiang2023review} and natural language processing~\cite{bharadiya2023comprehensive,khurana2023natural}, often surpassing human performance in many cases. However, studies have shown that these high-performance models are extremely sensitive to small changes in input data, making them vulnerable to adversarial attacks~\cite{goodfellow2015explainingharnessingadversarialexamples}. These attacks involve subtle and almost imperceptible modifications to the input data, causing the model to produce incorrect outputs.

Therefore, studying adversarial attacks is crucial for understanding the vulnerabilities of deep learning models. By exploring how to generate adversarial samples that can deceive models, researchers can gain deeper insights into the decision boundaries and learning mechanisms of these models, as well as reveal the inner workings of the models~\cite{shaham2018understanding,wang2023better,zhang2024does}. These insights are essential for enhancing the robustness of models, making them more secure and reliable, and advancing the theory of deep learning.

Currently, adversarial attacks can be categorized into white-box and black-box attacks based on the amount of information available to the attacker~\cite{jin2024benchmarking}. White-box attacks assume the attacker has complete knowledge of the victim model, including its architecture, parameters, and training dataset~\cite{goodfellow2015explainingharnessingadversarialexamples}, while black-box attacks are conducted without any internal knowledge of the model, making them more representative of real-world scenarios~\cite{papernot2017practical}. Within black-box attacks, there are further subdivisions based on the attack initiation method: query-based attacks and transferability attacks~\cite{jin2023danaa}.

\begin{figure*}[t]
    \centering
    \includegraphics[width=\linewidth]{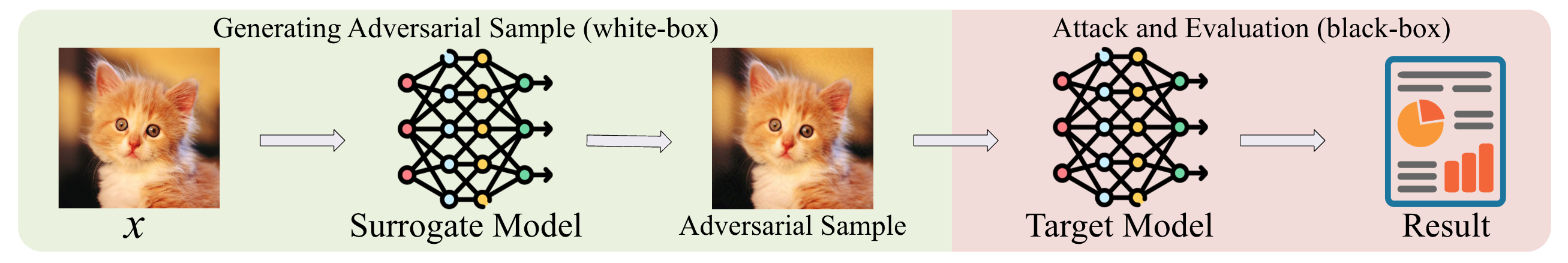}
    \caption{The flowchart of transferability-based attack}
    \label{fig.flowchart-tba}
\end{figure*}

Query-based attacks typically involve querying the model to simulate the victim model, and then using this simulated model to attack the victim model. However, this method has an inherent disadvantage as it requires frequent access to the target model, which reduces the stealthiness of the attack~\cite{jin2023danaa}. In this paper, we focus on the other type of black-box attack, namely transferability attacks. The principle behind this approach is to use a surrogate model to generate adversarial samples that are also effective against the victim model. Compared to query-based attacks, transferability attacks are more stealthy and pose a greater threat in real-world attack scenarios. Therefore, our primary focus in this paper is on transferability attack methods. As shown in Figure~\ref{fig.flowchart-tba}, the process of a transferability attack is illustrated. First, the original image is input into the source model, generating adversarial examples by attacking this source model. Subsequently, these examples are transferred to the target model, and their impact is evaluated again. This process demonstrates how adversarial examples crafted for one model can affect other black-box models without access to internal information, providing a clear illustration of the concept of transferability-based attack methods.

Existing transferability-based methods can be broadly classified into five categories based on their principles: Generative Architecture, Semantic Similarity, Gradient Editing, Target Modification, and Ensemble Approach~\cite{jin2024benchmarking}. Generative Architecture methods use Generative Adversarial Networks (GANs) to approximate the decision boundaries of the target model, allowing for the rapid generation of adversarial samples once the generator is trained. Semantic Similarity methods enhance attack transferability by constructing samples that are semantically similar to the original ones. Gradient Editing methods generate adversarial samples by modifying or optimizing gradient information, independent of the specific attack characteristics. Target Modification strategies exploit common features across different models for attacks. Lastly, the Ensemble Approach combines multiple models to improve the transferability and robustness of the attack.

\begin{figure*}
    \centering
    \includegraphics[width=\linewidth]{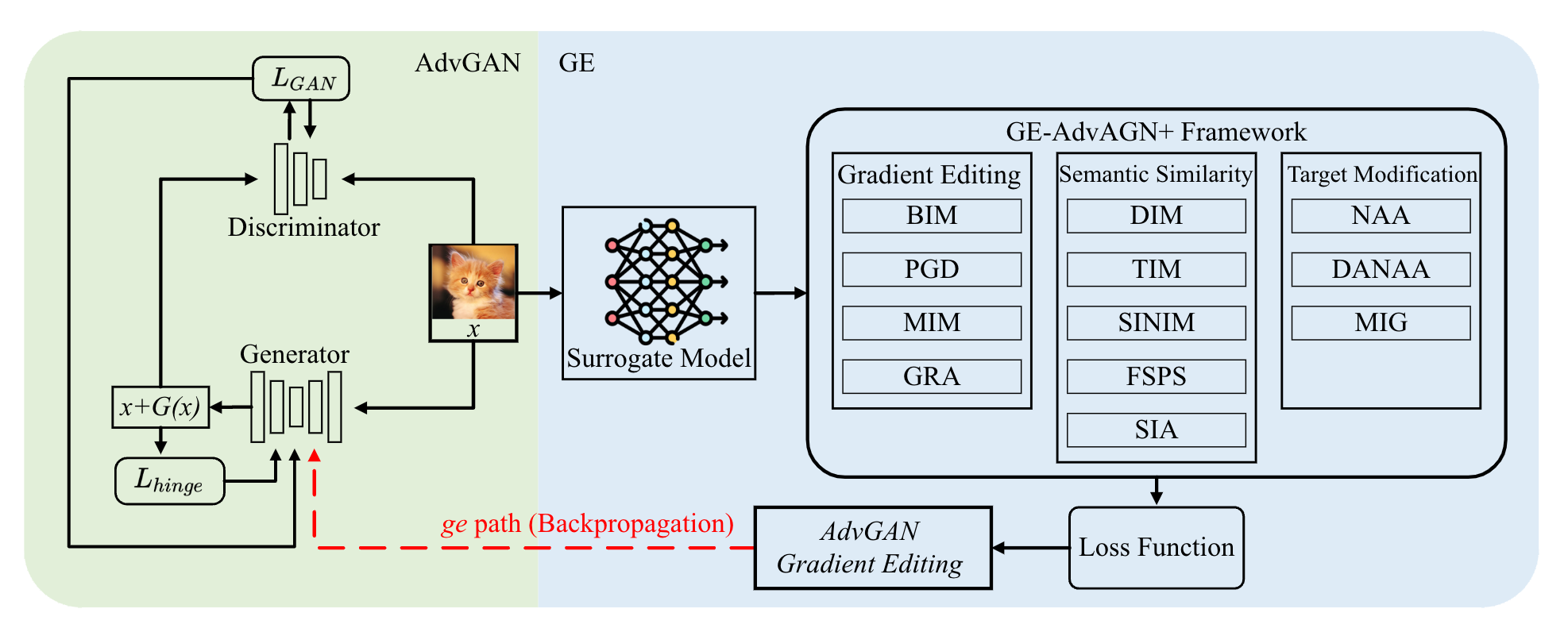}
    \caption{The flowchart of AdvGAN+ Framework}
    \label{fig.flowchart}
\end{figure*}

However, existing transferability-based adversarial attack methods still have certain limitations. For instance, some methods are slow in inference, which affects their feasibility in real-time applications such as edge computing. Although AdvGAN~\cite{xiao2018generating} addresses some of these issues, it does not fully leverage available information, leading to suboptimal transferability of the generated adversarial samples, as we discuss in detail in Section~\ref{sec.advgan}. To overcome these limitations, GE-AdvGAN is introduced, combining the principles of Generative Architecture and Gradient Editing. It uses frequency exploration as a basis for gradient editing to generate highly transferable adversarial samples~\cite{zhu2024ge}. However, there remains untapped potential within the model, such as internal feature representations and interactions between different layers. Therefore, as shown in Figure~\ref{fig.flowchart}we extend the application of gradient editing in GE-AdvGAN+ Framework by integrating more types of information and techniques, providing a more comprehensive exploration and optimization of the adversarial sample generation mechanism. This advancement not only improves the effectiveness of adversarial samples but also offers new perspectives for understanding the inherent vulnerabilities of deep learning models.

Specifically, in GE-AdvGAN++, we consider not only gradient information but also intermediate layer features within the model and similarities across different models to further enhance the transferability of adversarial samples. Additionally, compared to traditional transferability-based adversarial attack methods, GE-AdvGAN++ significantly improves computational efficiency. We also explore how fine-grained target gradient modification strategies can selectively attack key features of the model, thereby significantly improving attack effectiveness without incurring additional computational costs. Extensive experiments demonstrate the performance of GE-AdvGAN++ across multiple standard datasets and models, showing significant advantages in improving adversarial sample transferability and attack success rates while maintaining low computational complexity.

Our contributions can be summarized as follows:

\begin{enumerate}
    \item We provide an in-depth analysis and discussion of the core principles of nearly all mainstream transferability-based attack methods and, for the first time, mathematically define these methods' classifications.
    \item Through rigorous mathematical proofs, we demonstrate the feasibility of combining these methods with GE-AdvGAN+ Framework, providing a new theoretical foundation for adversarial attack research.
    \item Extensive experiments across multiple standard datasets and deep learning models confirm the effectiveness of GE-AdvGAN++. These experiments reveal the great potential of gradient editing techniques and show significant advantages in enhancing adversarial sample transferability and attack success rates while ensuring computational efficiency.
    \item To promote further research and development in the academic community, we have decided to open-source the implementation code of GE-AdvGAN++. This open-source contribution will enable the research community to reproduce our results, verify our theories, and build upon them for further innovation and optimization.
\end{enumerate}

This study extends our previous conference paper~\cite{zhu2024ge}.
In Section~\ref{introduction}, we expanded the conceptual diagram of GE-AdvGAN~\cite{zhu2024ge} to help readers understand the GE-AdvGAN+ architecture more clearly and comprehensively. In Section~\ref{sec_related_work}, we provided a more comprehensive and systematic introduction to the classification, principles, advantages, and disadvantages of different types of transferable attacks, so that readers can gain a more accurate and complete understanding of the field. Section~\ref{sec_methodology} delves into the expandability of GE-AdvGAN and explores its compatibility by combining various types of gradient information with GE-AdvGAN. In Section~\ref{experiment}, we analyze the impact of different types of gradient information on the performance of GE-AdvGAN+.

In this study, we made the following innovations in expanding GE-AdvGAN:
\begin{itemize}
\item Encapsulated GE-AdvGAN into an extensible generative adversarial attack transferability framework called GE-AdvGAN+.
\item Verified the framework on more models, including ViT models in addition to traditional CNN models.
\item Adopted more complex and effective gradient editing techniques, utilizing up to 12 different types of gradient information pairs, making the generated adversarial samples more transferable.
\item Proposed the most transferable generative attack method to date, GE-AdvGAN++.
\end{itemize}

\section{Background}\label{sec_related_work}
In this section, we discuss the current state of adversarial attacks, focusing on their principles, advantages, and limitations, as summarized in Table~\ref{tab.allmethods}. We first introduce some classic white-box adversarial attack methods, followed by a detailed overview of black-box transferability-based adversarial attacks. Finally, we examine generative structure-based attack methods.

\begin{table}[h!]
\centering
\caption{Summary of Various Adversarial Attack Methods: This table categorizes and compares different adversarial attack methods based on their type, subclass, principles, advantages, and disadvantages.}
\label{tab.allmethods}
\setlength{\tabcolsep}{8pt}
\normalsize
\resizebox{\textwidth}{!}{
\begin{tabular}{@{}>{\centering}m{1.2cm} >{\centering}m{2.0cm} >{\centering}m{1.2cm} >{\centering}m{2cm} >{\centering}m{1.8cm} >{\centering\arraybackslash}m{6.2cm} >{\centering\arraybackslash}m{2.2cm} >{\centering\arraybackslash}m{2.5cm}@{}}
\toprule
\makecell{Name} & \makecell{Reference} & \makecell{Year} & \makecell{Attack Type} & \makecell{Subclass} & \makecell{Principle} & \makecell{Advantages} & \makecell{Disadvantages} \\ \midrule
\makecell{BIM} & \cite{kurakin2018adversarial} & ICLR 2017 & White-box & - & Iteratively perturbs input data in the direction of the gradient sign of the loss function to generate adversarial examples & Simple and easy to implement & Risk of overfitting, poor transferability \\ \midrule
\makecell{PGD} & \cite{madry2017towards} & ICLR 2018 & White-box & - & Introduce a random perturbation and iteratively generates adversarial examples, a generalization of BIM & Effective and simple & Prone to overfitting \\ \midrule
\makecell{AdvGAN} & \cite{xiao2018generating} & IJCAI 2018 & Black-box & Generative Architecture & Uses generator and discriminator to iteratively approach the optimal perturbations & Efficient, user-friendly, good transferability & Does not fully exploit model information \\ \midrule
\makecell{MI-FGSM} & \cite{dong2018boosting} & CVPR 2018 & Black-box & Gradient Editing & Incorporates momentum to accumulate past gradient information for updating perturbations & Enhances transferability, avoids local optima & Potential for excessive optimization \\ \midrule
\makecell{TI-FGSM} & \cite{dong2019evading} & CVPR 2019 & Black-box & Semantic Similarity & Applies multiple random translations to the input image, then averages the gradients to generate translation-invariant perturbations & Efficient, versatile, combinable with other attacks & Additional computational cost, parameter tuning needed \\ \midrule
\makecell{DI-FGSM} & \cite{xie2019improving} & CVPR 2019 & Black-box & Semantic Similarity & Applies random transformations to the input image before calculating gradients and updating adversarial examples & Enhanced transferability, simple and effective & Dependent on input transformations \\ \midrule
\makecell{SINI-FGSM} & \cite{lin2019nesterov} & ICLR 2020 & Black-box & Semantic Similarity & Combines Scale-Invariance and Nesterov Accelerated Gradient to improve transferability and efficiency & High transferability and efficiency & Requires extensive parameter tuning \\ \midrule
\makecell{FIA} & \cite{wang2021feature} & ICCV 2021 & Black-box & Target Modification & Considers the importance of model features to guide perturbation generation & High attack efficiency, wide applicability & Additional computational overhead \\ \midrule
\makecell{NAA} & \cite{zhang2022improving} & CVPR 2022 & Black-box & Target Modification & Targets neurons contributing most to the decision process for more effective transferable adversarial examples & Breaks through defense methods & Complexity, extensive parameter tuning needed \\ \midrule
\makecell{DANAA} & \cite{jin2023danaa} & ADMA 2023 & Black-box & Target Modification & Uses adversarial non-linear attribution paths for generating transferable adversarial examples & Accurate attribution, improved transferability & Increased computational complexity \\ \midrule
\makecell{FSPS} & \cite{zhu2023improving} & CIKM 2023 & Black-box & Semantic Similarity & Explores stationary points on the loss curve as starting points, using frequency search for attack direction & Superior transferability with exploration strategies & Increased computational cost \\ \midrule
\makecell{GRA} & \cite{zhu2023boosting} & ICCV 2023 & Black-box & Gradient Editing & Identifies the relevance of input image gradients and establishes their relationship through cosine similarity & Adaptive correction of update direction & Increased computational complexity on large models \\ \midrule
\makecell{MIG} & \cite{ma2023transferable} & ICCV 2023 & Black-box & Target Modification & Uses integrated gradients, considering the accumulation of past gradients in a momentum-like manner & High success rate and transferability & Computational complexity, depends on assumptions \\ \midrule
\makecell{SIA} & \cite{wang2023structure} & ICCV 2023 & Black-box & Semantic Similarity & Applies different transformations locally in various regions of the input image for more diverse transformed images & Preserves global structure, maintains semantic information & Increased computational cost \\ \midrule
\makecell{GE-AdvGAN} & \cite{zhu2024ge} & SDM 2024 & Black-box & Generative Architecture; Gradient Editing & Uses Discrete Cosine Transformation to map samples to frequency domain for gradient editing direction & Increases attack efficiency, effective across architectures & Focuses only on frequency domain, time-consuming training \\ \bottomrule
\end{tabular}
}
\end{table}

\subsection{White-box Adversarial Attacks}

White-box attacks are crucial in the study of adversarial attacks as they allow attackers full knowledge of the target model's architecture and parameters. Notable white-box attack methods include the Fast Gradient Sign Method (FGSM)~\cite{goodfellow2015explainingharnessingadversarialexamples} and its iterative version, the Basic Iterative Method (BIM)~\cite{kurakin2018adversarial}. FGSM constructs perturbations by leveraging the gradients of the loss function with respect to the input data, offering a fast but sometimes suboptimal attack. To enhance attack effectiveness, the BIM method was proposed, which iteratively adjusts the input data along the direction of the gradient sign, thereby increasing the attack success rate. Following FGSM and BIM, Madry et al~\cite{madry2017towards}. introduced the Projected Gradient Descent (PGD) attack, a more robust and powerful white-box method. PGD can be seen as an extension of BIM, performing a more exhaustive search within an $\epsilon$-ball around the original input, addressing a constrained optimization problem aimed at maximizing model loss. Additionally, the C\&W~\cite{carlini2017towards} attack generates minimal and effective perturbations by constructing and optimizing a new objective function, which is also utilized in AdvGAN~\cite{xiao2018generating}. These white-box methods have laid the foundation for subsequent research in adversarial attacks and defenses.

\subsection{Black-box Adversarial Attacks}

In the study of black-box adversarial attacks, transferability-based methods can be categorized into five main types based on their principles~\cite{jin2024benchmarking}. The first type is generative architecture methods, which use Generative Adversarial Networks (GANs)~\cite{goodfellow2020generative} or similar generative models to create adversarial samples transferable across different models, emphasizing the rapid generation of new adversarial samples once the generator is trained. The second type, semantic similarity methods, focuses on maintaining semantic consistency by attacking samples semantically related to the original ones, thereby extending the transferability of adversarial samples. Gradient editing methods generate more effective adversarial samples on target models by analyzing and adjusting gradient information, making the generated samples more stealthy and transferable. Target modification methods exploit similar features across different models, such as model interpretability similarities, to achieve adversarial sample transfer by directly attacking these similar features, thus enhancing transferability. Lastly, ensemble methods combine feedback from multiple models to generate adversarial samples, improving transferability. However, acquiring suitable multiple surrogate models is challenging in practice, and comparing multiple surrogate models with a single model is unfair. Additionally, using multiple models consumes more computational resources and time, so this category is not within the scope of our research.

\subsubsection{Gradient Editing}

In the domain of adversarial attacks, gradient editing techniques have emerged as new strategies to improve the precision and stealth of attacks. The Momentum Iterative Fast Gradient Sign Method (MI-FGSM)~\cite{dong2018boosting} and Gradient Relevance Attack (GRA)~\cite{zhu2023boosting} are two prominent examples of this approach.

MI-FGSM~\cite{dong2018boosting} is an enhancement of the Basic Iterative Method (BIM). It accumulates the gradient information from previous steps in each iteration, thereby enhancing the transferability of the adversarial samples. This momentum accumulation method improves the robustness of the attack, making the generated adversarial samples more transferable across different models, and thus harder for model defenses to detect and resist. The advantage of MI-FGSM is its improved transferability and ability to avoid local optima during attacks through the momentum mechanism. However, in some cases, over-reliance on historical gradients may lead to adversarial samples that are overly optimized for the source model, potentially reducing their effectiveness on target models.

On the other hand, GRA~\cite{zhu2023boosting} adopts a different perspective, utilizing the correlation of input image gradients to adaptively adjust the update direction. GRA identifies the correlation among the gradients of input images, treating the current gradient as a query vector and neighboring gradients as key vectors, establishing a relationship through cosine similarity. This method can adaptively determine the update direction based on the gradient neighborhood information, thus generating more effective adversarial samples. The main advantage of GRA is its adaptive correction of the update direction using neighborhood gradient information, increasing the effectiveness of generated adversarial samples. However, the drawback is increased computational complexity, especially when handling large models and complex datasets.

\subsubsection{Semantic Similarity}

In the field of adversarial attacks, research on semantic similarity focuses on generating adversarial samples that can deceive deep learning models while preserving the semantic content of the input data. This type of method aims to create adversarial samples that are virtually indistinguishable from the original samples to human observers but can cause deep learning models to make incorrect predictions. Key methods in this domain include Translation-Invariant FGSM (TI-FGSM)~\cite{dong2019evading}, Diversity Input FGSM (DI-FGSM)~\cite{xie2019improving}, Scale-Invariant Nesterov Iterative FGSM (SI-NI-FGSM)~\cite{lin2019nesterov}, Frequency-based Stationary Point Search (FSPS)~\cite{zhu2023improving}, and Structure-Invariant Attack (SIA)~\cite{wang2023structure}.

TI-FGSM~\cite{dong2019evading} improves attack stealth and effectiveness by enhancing the translation invariance of images. In each iteration, TI-FGSM performs multiple random translations of the image, computes gradients on these translated images, and synthesizes these gradients to generate translation-invariant adversarial perturbations. This approach ensures that the adversarial samples maintain their effectiveness even after spatial transformations, such as translations, by the model, enhancing the adaptability and stealth of the samples. However, this method increases computational cost due to the need for multiple translations and gradient computations compared to traditional FGSM methods.

DI-FGSM~\cite{xie2019improving} enhances the transferability of adversarial samples by introducing input diversity. In each iteration, DI-FGSM applies random transformations, such as scaling and cropping, to the original image and computes gradients on these transformed images to update the adversarial sample. This method generates adversarial samples that are effective across different models and preprocessing steps, enhancing sample generalization and stealth. However, the effectiveness of DI-FGSM heavily depends on the choice of input transformations and parameter settings, requiring careful design for optimal results.

SI-NI-FGSM~\cite{lin2019nesterov} combines scale invariance and Nesterov accelerated gradient descent to improve the transferability and efficiency of adversarial samples. By performing multi-scale transformations on the original image and averaging the gradients, SI-NI-FGSM ensures that the generated adversarial samples maintain their effectiveness across different scale transformations, demonstrating high robustness and stealth. The use of Nesterov acceleration also makes the adversarial sample generation process more efficient, though the multiple parameter adjustments involved add complexity.

FSPS~\cite{zhu2023improving} explores stationary points on the loss curve as starting points for attacks and optimizes the attack direction through frequency domain search. This method aims to find the optimal local solutions to improve attack effectiveness and adaptability. The FSPS method is characterized by its ability to quickly locate efficient adversarial sample generation paths, though this frequency domain exploration-based approach is relatively computationally expensive and dependent on the model's frequency response characteristics.

SIA~\cite{wang2023structure} applies different transformations to local regions of the input image, aiming to increase sample diversity while maintaining the overall structure of the image, thus enhancing the stealth and effectiveness of adversarial samples. This method focuses on introducing effective perturbations while preserving semantic information, representing an innovative attempt in adversarial sample generation strategies. However, the challenge of SIA lies in balancing the degree of local transformations and maintaining global semantics, as well as the associated computational cost.

\subsubsection{Target Modification}

Target modification techniques in the field of adversarial attacks focus on generating more precise and difficult-to-detect adversarial samples by meticulously analyzing and adjusting attack strategies. These methods pay special attention to the importance of specific features or neurons within the model, guiding the generation of adversarial perturbations based on these factors to enhance attack effectiveness and transferability. In this area, methods such as Feature Importance-Aware Attack (FIA)~\cite{wang2021feature}, Neuron Attribution-based Attack (NAA)~\cite{zhang2022improving}, Double Adversarial Neuron Attribution Attack (DANAA)~\cite{jin2023danaa}, and Momentum Integrated Gradients (MIG)~\cite{ma2023transferable} demonstrate the potential of target modification to enhance adversarial attack performance.

The FIA method~\cite{wang2021feature} is a feature importance-based attack strategy that selectively applies perturbations to features that significantly contribute to the model's decision-making process. This strategy not only improves attack efficiency but also enhances the transferability of adversarial samples across different models. However, accurately assessing feature importance may come with high computational costs, especially when dealing with complex models.

The NAA method~\cite{zhang2022improving} delves deeper into the model's internal structure, particularly neuron attribution, to generate more targeted adversarial samples. NAA focuses on the most critical parts of the model's decision-making process, resulting in adversarial samples with better transferability across multiple models. However, the computational complexity of this method is relatively high, especially when fine-grained analysis of the model's internal structure is required.

DANAA~\cite{jin2023danaa} further extends neuron attribution-based attack methods by employing a double attribution mechanism and using nonlinear paths to evaluate neuron importance, generating more effective adversarial samples. This method provides new perspectives for understanding and leveraging the internal mechanisms of deep learning models, but its implementation complexity also increases, potentially requiring more computational resources and time to generate adversarial samples.

MIG~\cite{ma2023transferable} employs an integrated gradients-based approach, emphasizing the similarity of gradients across different models and optimizing the generation of adversarial perturbations through momentum-like accumulation of previous and current gradients. This method enhances the transferability and success rate of adversarial samples, although the computation of integrated gradients may increase the complexity of the attack process.

\section{Methods}\label{sec_methodology}

\subsection{Definition of Adversarial Attack}

An adversarial attack can be defined as an optimization problem where the goal is to find the smallest perturbation $\delta$ such that the model $f$ misclassifies the perturbed input $x + \delta$, compared to the original input $x$. Specifically, the objective is to identify a minimal $\delta$ such that $f(x + \delta) \neq y$, where $y = f(x)$ represents the original label predicted by the model for the input $x$. The mathematical formulation is as follows:
\begin{align*}
\text{minimize} \quad & \| \delta \|_p \\
\text{subject to} \quad & f(x + \delta) \neq f(x) \\
& x + \delta \in \mathcal{X}
\end{align*}
where $x$ is the original input, $\delta$ is the perturbation added to $x$, $f$ is the model under attack, and $\mathcal{X}$ denotes the valid input space. The term $\| \delta \|_p$ represents the $p$-norm of $\delta$, with common choices for $p$ being 1, 2, or $\infty$. The choice of $p$ affects the sparsity or size of the perturbation, ensuring that the perturbation is minimal while still causing the model to produce incorrect predictions.

\subsection{Mathematical Definition of AdvGAN}\label{sec.advgan}

AdvGAN~\cite{xiao2018generating} aims to train a generative model $G$ that outputs a perturbation $\delta$ such that the perturbed input $x + \delta$ can deceive the classifier $f$. During training, both the generator $G$ and a discriminator $D$ are optimized, where $D$ is tasked with distinguishing real samples from adversarial samples generated by $G$. The optimization objective can be expressed as the following minimax problem:
\begin{align*}
\min_G \max_D \quad & \mathbb{E}_{x \sim X} \left[ \log D(x) + \log (1 - D(G(x))) \right] \\
\text{subject to} \quad & f(x + G(x)) \neq f(x) \\
& x + G(x) \in \mathcal{X}
\end{align*}
Additionally, to ensure that the generated perturbation is both minimal and effective, an extra loss function is often included to constrain $G$:
\begin{align*}
\mathcal{L}_{adv}(G) &= \alpha \cdot \mathbb{E}_{x \sim X} \|G(x)\| - \beta \cdot \mathbb{E}_{x \sim X} \mathcal{L}(f(x+G(x)), y)
\end{align*}
where $\mathcal{L}$ is typically a cross-entropy loss function measuring the difference between $f(x+G(x))$ and the target label $y$. The parameters $\alpha$ and $\beta$ are hyperparameters that balance the perturbation size and the adversarial success rate. The goal of AdvGAN is to generate perturbations that are both hard to detect and effective in misleading the classifier $f$. Essentially, AdvGAN can be understood as integrating the training process of C\&W~\cite{carlini2017towards} into the training of the generator, although C\&W's poor transferability affects AdvGAN's transferability performance.

\subsection{GE-AdvGAN}

GE-AdvGAN~\cite{zhu2024ge} modifies the gradient update mechanism of the generator $G$ to adjust the adversarial perturbation $\delta = G(x)$ it generates. This approach involves two key techniques: frequency domain exploration and gradient editing.

In the frequency domain exploration step, the input sample $x$ and the generated perturbation $G(x)$ are transformed into the frequency domain, typically using the Discrete Cosine Transform (DCT)~\cite{jia2022exploring}. This transformation allows for the analysis and modification of characteristics at different frequencies. By adjusting the amplitude and phase of the frequency components, a series of approximate samples $x_{f_i}$ are generated, which are used in training to further optimize the generator $G$.

The gradient editing process targets the parameter updates of the generator $G$. Specific gradient editing techniques, as shown in the following expression, are used to directly modify the gradients affecting the updates of $G$'s parameters:
\begin{equation}
    ge : \frac{\partial L}{\partial (x+G(x))}
\end{equation}
\begin{equation}
\begin{aligned}
W_{G} &= W_{G} - \eta \left( ge \cdot \frac{\partial (x + G(x))}{\partial G(x)} \cdot \frac{\partial G(x)}{\partial W_{G}} + \alpha \frac{\partial L_{GAN}}{\partial W_{G}} + \beta \frac{\partial L_{hinge}}{\partial W_{G}} \right)
\end{aligned}
\end{equation}

where $ge$ is the target gradient calculated based on the frequency domain analysis results, replacing the original gradient calculation $\nabla_{W_{G}} L_{adv}^{f}$.

To implement GE-AdvGAN, the sample $x$ is first transformed into the frequency domain using Discrete Cosine Transform (DCT), and variant samples $x_{f_i}$ are generated based on different frequency sensitivities. These samples are then used to compute the target gradient $ge$, which feeds back into the training process of the generator $G$. The gradient $ge$ is the average of gradients corresponding to multiple $x_{f_i}$, optimizing the generation of adversarial samples. Notably, in this study, the $ge$ component can be replaced with any other gradient information, and apart from the frequency domain, more dimensions of information can be explored.

\subsection{Gradient Editing with Additional Information}

As previously mentioned, gradient editing can modify the direction of perturbation learned by the generator, thereby enhancing the transferability of generated adversarial samples. Currently, there are various methods for gradient editing, according to~\cite{jin2024benchmarking}, including five different types of adversarial attacks: Generative Architecture, Semantic Similarity, Gradient Editing, Target Modification, and Ensemble Approach.

Our experiments reveal that gradients from the Semantic Similarity category are more easily learned by the generator. As discussed earlier, the focus of this paper is on the Generative Architecture type of attacks due to their computational speed and efficiency. The Ensemble Approach, on the other hand, is challenging to deploy in real-world scenarios as it requires multiple surrogate models, which are difficult to obtain. Moreover, using multiple surrogate models is unfair compared to single-model methods.

In this paper, we integrate Gradient Editing, Semantic Similarity, and Target Modification methods with our previously developed GE-AdvGAN framework. We explore the scalability of GE-AdvGAN by incorporating more gradient editing techniques, enhancing the framework's capability to edit gradients. The following sections provide the mathematical definitions of Gradient Editing, Semantic Similarity, and Target Modification methods, with gradient editing techniques designated as $GE_{(\cdot)}$. For clarity, we use $g^{(k+1)}$ to denote components in the equations that can be incorporated into the GE-AdvGAN gradient editing framework. These components allow for the generation of adversarial samples using the following formula:

\begin{equation}
x_{f}^{(k+1)} = \text{Clip}_{x_{0}, \epsilon} \left( x_{f}^{(k)} + \alpha \cdot \text{sign} \left( g^{(k+1)} \right) \right)
\end{equation}

\subsubsection{Mathematical Definition of Gradient Editing}
Gradient Editing type adversarial attacks primarily optimize the objective function by adjusting the gradient $\nabla_x L(f(x), y)$ with respect to the input $x$ to generate adversarial samples. The editable part of such methods can be mathematically expressed as follows:

\begin{equation}
ge = \nabla_x L(f(x^{(k)}), y)
\end{equation}

\paragraph{$GE_{BIM}$ \& $GE_{PGD}$}

First, we combined the classical Gradient Editing methods BIM and PGD with our GE-AdvGAN+ framework. Although their mathematical principles are essentially the same, their initial inputs $x_{f}^{(0)}$ differ; in BIM, $x_{f}^{(0)}=x$, starting from the original sample $x$, while PGD starts from a point $x_{f}^{(0)}=x+Uniform(-\epsilon, \epsilon)$, with random perturbations around the original sample. Furthermore, PGD restricts each gradient update within a norm ball to ensure the perturbation does not exceed the preset maximum range, as shown in Equation~\ref{eq.pgd}. The gradient information for both methods is obtained via $\nabla_x L(f(x_{f}^{(k)}), y)$, which integrates with our framework's editing method.

\begin{equation}
\label{eq.pgd}
\begin{aligned}
g^{(k+1)}  &= \nabla_x L(f(x_{f}^{(k)}), y)  \\
x_{f}^{(k+1)} &= \text{Clip}_{x, \epsilon} \left( x_{f}^{(k)} + \alpha \cdot \text{sign} \left( g^{(k+1)}  \right) \right)
\end{aligned}
\end{equation}

\paragraph{$GE_{MIM}$}

Next, we incorporated MIM~\cite{dong2018boosting} into our framework. In Equation~\ref{eq.mim}, $g^{(k)}$ represents the accumulated gradient (momentum term) from the $k$-th iteration. The momentum factor $\mu$, typically set between 0.9 and 1.0, balances the influence of previous and current gradients. The $g^{(k+1)}$ can be integrated with the GE-AdvGAN framework.

\begin{equation}
\label{eq.mim}
\begin{aligned}
g^{(0)} &= 0 \\
g^{(k+1)}  &= \mu \cdot g^{(k)} + \frac{\nabla_x L(f(x_{f}^{(k)}), y)}{\| \nabla_x L(f(x_{f}^{(k)}), y) \|_1} \\
\end{aligned}
\end{equation}

\paragraph{$GE_{GRA}$}

We also integrated GRA~\cite{zhu2023boosting} into our framework. In Equation~\ref{eq.gra}, $G$ represents the gradient of the current input, while $G_i$ denotes the gradient of the $i$-th neighboring sample. The weights $w_i$ are calculated based on the cosine similarity between $G$ and $G_i$ using a softmax function, where $\tau$ is the temperature parameter that controls the smoothness of the softmax function, $\text{Relevance}(G, x_{f}^{(k)}, y)$ represents the update direction weighted based on gradient and neighborhood information at step $k$. Our GE-AdvGAN method can leverage the gradient information $g^{(k+1)}$ to enhance the transferability of generated samples.

\begin{equation}
\label{eq.gra}
\begin{aligned}
w_i &= \frac{\exp(\cos(G, G_i) / \tau)}{\sum_{j=1}^{n} \exp(\cos(G, G_j) / \tau)} \\
\text{Relevance}&(G, x_{f}^{(k)}, y) = \sum_{i=1}^{n} w_i \cdot G_i \\
g^{(k+1)}  &= \mu \cdot g^{(k)} + \frac{\text{Relevance}(G, x_{f}^{(k)}, y)}{\|\text{Relevance}(G, x_{f}^{(k)}, y)\|_1} \\
\end{aligned}
\end{equation}

\subsubsection{Semantic Similarity}
Semantic Similarity-based adversarial attacks focus on enhancing transferability by altering input data through data augmentation techniques to construct samples semantically similar to the original data. The part where gradient editing can be performed is mathematically defined as follows:
\begin{equation}
ge = \nabla_x L(f(D_{\rho}(x^{(k)})), y) 
\end{equation}

where $D_{\rho}$ represents a series of parameterized data augmentation operations such as random scaling, cropping, and translation.

\subsubsection{$GE_{DIM}$}

In Semantic Similarity-based adversarial attacks, we first combined the typical DIM~\cite{xie2019improving} method with our approach. In Equation~\ref{eq.dim}, $D_\rho$ denotes random transformations applied to the input $x_{f}^{(k)}$, such as random scaling and cropping, which aim to minimally affect the original image semantics, with $\rho$ as the transformation parameter. The $g^{(k+1)}$ part can be integrated with the GE-AdvGAN+ framework.

\begin{equation}
\label{eq.dim}
\begin{aligned}
g^{(k+1)}  &= \mathbb{E}_{\rho} [\nabla_x L(f(D_\rho(x_{f}^{(k)})), y)]\\
\end{aligned}
\end{equation}

\paragraph{$GE_{TIM}$}

TIM~\cite{dong2019evading} can also be incorporated into our framework. In Equation~\ref{eq.tim}, $T_\tau$ denotes the translation transformation (a geometric operation that shifts an image from one location to another without altering its orientation, size, or shape) applied to the input image $x_{f}^{(k)}$, with $\tau$ as the translation parameter. $\mathbb{E}_{\tau}$ denotes the expectation over all possible translations $\tau$, averaging the gradients across multiple translations.

\begin{equation}
\label{eq.tim}
\begin{aligned}
g^{(k+1)}  &= \mathbb{E}_{\tau} [\nabla_x L(f(T_\tau(x_{f}^{(k)})), y)]\\
\end{aligned}
\end{equation}

\paragraph{$GE_{SINIM}$}

Next, we integrated SINIM~\cite{lin2019nesterov} with our framework. In Equation~\ref{eq.sinim}, $N(0, \sigma)$ denotes Gaussian noise with a mean of 0 and standard deviation $\sigma$, injected into the adversarial sample $x_{f}^{(k)}$ at each iteration. This noise does not affect the semantic meaning of the original image. The $g^{(k+1)}$ component fits well with our GE-AdvGAN+ framework.

\begin{equation}
\label{eq.sinim}
\begin{aligned}
g^{(k+1)}  &=\mathbb{E}_{\sigma} [\nabla_x L(f(x_{f}^{(k)} + N(0, \sigma)), y)] \\
\end{aligned}
\end{equation}

\paragraph{$GE_{FSPS}$}

In this section, we integrate FSPS~\cite{zhu2023improving} with the GE-AdvGAN+ framework, denoting this gradient editing method as $GE_{FSPS}$. In Equation~\ref{eq.fsps}, $k$ represents the current iteration step, and $w$ is a predetermined threshold for a warm-up step, determining when to transition from the adversarial sample to a stationary point as the attack starting point. $x[k<w]$ indicates using the current adversarial sample $x$ before the warm-up step $w$, while $sp[t \geq w]$ indicates using the stationary point $sp$ after the warm-up step $w$. The gradient information $g^{(k+1)}$ integrates with our GE-AdvGAN+ framework. Notably, after applying FSPS's gradient information computation method, the transferability of adversarial attacks learned by the GE-AdvGAN+ framework is strongest, leading us to designate it as GE-AdvGAN++.

\begin{equation}
\label{eq.fsps}
\begin{aligned}
x_{dct} &= DCT(x[k<w] + sp[k \geq w] + N(0, I) \cdot \frac{\epsilon}{255}) \\
x_{idct} &= IDCT(x_{dct} \ast N(1, \sigma)) \\
g^{(k+1)} &= \mathbb{E} \left[ \nabla_x L(x_{idct}, y) \right]
\end{aligned}
\end{equation}

\paragraph{$GE_{SIA}$}

In this section, we incorporate SIA~\cite{wang2023structure} into the GE-AdvGAN+ framework, referring to this gradient editing method as $GE_{SIA}$. In Equation~\ref{eq.sia}, $N$ represents the number of transformed images, and $T_n(\cdot)$ denotes the $n$-th transformation applied to the input $x$. The $g^{(k+1)}$ component can be integrated with the GE-AdvGAN+ framework, enhancing the transferability of adversarial samples generated by the generator.

\begin{equation}
\label{eq.sia}
\begin{aligned}
g^{(k+1)}  &= \mathbb{E}_{T_n} [\nabla_x L(f(T_n(x_{f}^{(t)})), y)]\\
\end{aligned}
\end{equation}

\subsubsection{Target Modification}
This type of method assumes that most models focus on similar features of the samples. By using interpretability methods such as attribution, the important features that the surrogate model focuses on are identified and then attacked to improve performance when targeting the victim model. The editable part of such methods can be mathematically expressed as follows:

\begin{equation}
ge = \nabla_x L(A(f(x^{(k)})), y)
\end{equation}

where $L(\cdot)$ will be modified for a specific target, $A(\cdot)$ represents interpretability methods, such as~\cite{sundararajan2017axiomatic,pan2021explaining,zhu2024iterative,zhu2023mfaba,zhu2024attexplore}. More detailed formulas for interpretability methods can be found in the appendix.

\paragraph{$GE_{NAA}$}

In this section, we first integrate the NAA~\cite{zhang2022improving} method from Target Modification into our framework. The NAA method is based on the assumption that attribution results from different models are likely to be similar. It modifies the target from attacking the loss function to attacking the attribution results. In Equation~\ref{eq.naa}, $n$ represents the number of integrated steps, $A_y$ is the sum of the attribution results on the target layer $y$, i.e., the sum of all neuron attribution values, $IA(\phi_j)$ is Integrated Attention for latent feature $\phi_j$, computed by summing the gradients along the path integral, $N$ is the Neural Network Model, $W_{A_y}$ represents weighted attribution, $f_p$ and $f_n$ are functions applied to positive and negative attributions, respectively, and $\beta$ is a hyperparameter balancing positive and negative attributions. The $\theta$ represents intermediate layer features. The $g^{(k+1)}$ component can be incorporated into our GE-AdvAGN method.

\begin{equation}
\label{eq.naa}
\begin{aligned}
 A_\phi &= \sum_{\phi_j \in \phi} A_{\phi_j} = \sum_{\phi_j \in \phi} \Delta \phi_j \cdot IA(\phi_j) \\
 IA(\phi_j) &= \frac{1}{n} \sum_{m=1}^n \left( \frac{\partial N}{\partial \phi_j}(\phi(x_m)) \right) \\
 W_{A_\phi} &= \sum_{A_{\phi_j} \geq 0} f_p(A_{\phi_j}) - \beta \cdot \sum{A_{\phi_j} < 0} f_n(-A_{\phi_j}) \\
 g^{(k+1)} &= \mu \cdot g^{(k)} + \frac{\nabla_x W_{A_\phi}}{\|\nabla_x W_{A_\phi}\|_1}
\end{aligned}
\end{equation}

\paragraph{$GE_{DANAA}$}

In this section, we integrate DANAA~\cite{jin2023danaa} into the framework. Compared to NAA, DANAA uses a nonlinear exploration path. In Equation~\ref{eq.danaa}, $N$ is the Neural Network Model, and $\gamma(y_j)$ represents the gradient in the nonlinear path. The $g^{(k+1)}$ component can also be applied to the GE-AdvGAN+ framework.

\begin{equation}
\label{eq.danaa}
\begin{aligned}
A_{y_j} &= \sum_{i=1}^{n^2} \int \Delta x_i \frac{\partial N(x_t)}{\partial y_j(x_t)} \frac{\partial y_j(x_t)}{\partial x_i^t} dt = \Delta y_j \cdot \gamma(y_j)\\
\gamma(y_j) &= \int \frac{\partial N(x_t)}{\partial y_j(x_t)} dt\\
W_{A_y} &= \sum_{A_{y_j} \geq 0} f_p(A_{y_j}) - \beta \cdot \sum_{A_{y_j} < 0} f_n(-A_{y_j}) \\
g^{(k+1)} &= \mu \cdot g^{(k)} + \frac{\nabla_x W_{A_y}}{\left\|\nabla_x W_{A_y}\right\|_1}\\
\end{aligned}
\end{equation}

\paragraph{$GE_{MIG}$}

In this section, we apply the gradients generated by the MIG~\cite{ma2023transferable} method to our framework. In Equation~\ref{eq.mig}, $b_k$ is the baseline input (usually a black image). $IG(\cdot)$ represents integrated gradients. The $g^{(k+1)}$ component can be combined with GE-AdvGAN to increase the transferability of the generated adversarial samples.

\begin{equation}
\label{eq.mig}
\begin{aligned}
IG(f, x_k, b_k) &= (x_k - b_k) \times \int_{\xi=0}^{1} \frac{\partial f(b_k + \xi \times (x_k - b_k))}{\partial x_k} d\xi \\
g^{(k+1)}  &= \mu \cdot g_{t} + \frac{IG(f,x_{f}^{(k)},y)}{\|IG(f,x_{f}^{(k)},y)\|_1} \\
\end{aligned}
\end{equation}

In summary, gradient editing with additional information can generate stronger and more robust adversarial samples by integrating additional information and features. These methods significantly improve the success rate of adversarial attacks and the diversity of adversarial samples by introducing techniques such as frequency domain, momentum, random transformations, feature masking, and nonlinear exploration into the generation process. We will conduct experimental analyses in Section~\ref{exp.geframe}.

\section{Experiments}\label{experiment}
\subsection{Experimental Setup}
\subsubsection{Datasets} 
To ensure fair comparison across various adversarial attack methods, this study adheres to the datasets used in prior works such as AdvGAN~\cite{xiao2018generating} and GE-AdvGAN~\cite{zhu2024ge}, as well as other methods like NAA~\cite{zhang2022improving}, SSA~\cite{long2022frequency}, DANAA~\cite{jin2023danaa}, and DANAA++~\cite{zhu2024rethinking}. The dataset comprises 1,000 randomly selected images from the ILSVRC 2012 validation set~\cite{russakovsky2015imagenet}, covering all 1,000 categories.

\subsubsection{Models} 
In our experiments, we utilized the four widely used models in image classification tasks as adopted in the AdvGAN paper: Inception-v3 (Inc-v3)~\cite{szegedy2016rethinking}, Inception-v4 (Inc-v4)~\cite{szegedy2017inception}, Inception-ResNet-v2 (Inc-Res-v2)~\cite{szegedy2017inception}, and ResNet-v2-152 (Res-152)~\cite{he2016deep} as source models. Additionally, we included the more complex ViT model MaxViT-T~\cite{tu2022maxvit} as a surrogate model to evaluate the generalizability of the GE-AdvGAN method. Furthermore, we tested the transferability of our method across seven models, including four standard training models (Inc-v3, Inc-v4, Inc-Res-v2, Res-152), and three adversarially trained models (Inc-v3-adv-ens3 ~\cite{tramer2017ensemble}, Inc-v3-adv-ens4~\cite{tramer2017ensemble}, and Inc-Res-adv-ens~\cite{tramer2017ensemble}).

\subsubsection{Metrics} 

We used the Adversarial Success Rate (ASR) as the primary metric for evaluation. ASR measures the effectiveness of adversarial attacks in altering the model's output. Specifically, it represents the proportion of samples where the classification result is successfully altered by the adversarial attack out of all tested samples. The formula for ASR is:

\begin{equation} 
ASR = \frac{N_{\text{successful attacks}}}{N_{\text{total samples}}} 
\end{equation}

where $N_{\text{successful attacks}}$ denotes the number of samples whose classification results were successfully altered after the adversarial attack, and $N_{\text{total samples}}$ is the total number of test samples. The ASR ranges from 0 to 1, with higher values indicating more significant effectiveness of the adversarial attack, demonstrating the model's susceptibility to being attacked.

\subsubsection{Baseline}

In this paper, we selected AdvGAN~\cite{xiao2018generating} as our baseline and GE-AdvGAN~\cite{zhu2024ge} as our main competing algorithm. Additionally, we compared the performance of 12 representative attack methods from the Gradient Editing, Semantic Similarity, and Target Modification categories, incorporating them into our GE-AdvGAN+ framework to evaluate their combined performance against their original implementations. The details of these 12 methods are listed in Table~\ref{tab:attackmethods}.

\begin{table}[]
\caption{Categorization of Adversarial Attack Methods Used in Experiments}
\label{tab:attackmethods}
\resizebox{0.6\linewidth}{!}{%
\begin{tabular}{@{}ll@{}}
\toprule
Attack Type         & Methods                    \\ \midrule
Gradient Editing    & BIM~\cite{kurakin2018adversarial}, PGD~\cite{madry2017towards}, MIM~\cite{dong2018boosting}, GRA~\cite{zhu2023boosting}         \\
Semantic Similarity & DIM~\cite{xie2019improving}, TIM~\cite{dong2019evading}, SINIM~\cite{lin2019nesterov}, FSPS~\cite{zhu2023improving}, SIA~\cite{wang2023structure} \\
Target Modification & NAA~\cite{zhang2022improving}, DANAA~\cite{jin2023danaa}, MIG~\cite{ma2023transferable}            \\ \bottomrule
\end{tabular}%
}
\end{table}

\subsubsection{Parameter Settings}

We maintained consistency with the parameter settings used in GE-AdvGAN~\cite{zhu2024ge}. Specifically, for the models Inc-v3, Inc-v4, Inc-Res-v2, Res-152, and MaxViT-T, the parameters \textit{adv lambd} and $N$ were set to 10, $\epsilon$ was set to 16, the number of epochs was set to 60, the change threshold was set to [20, 40], the Discriminator ranges were set to [1, 1], and the Discriminator learning rate was set to [0.0001,0.0001]. When Inc-v3 was used as the surrogate model, $\sigma$ was set to 0.7; for the other four models, $\sigma$ was set to 0.5. When Inc-Res-v2 was used as the source model, the Generator ranges and Generator learning rate were set to [1,1] and [0.0001, 0.000001], respectively; for the other four models, these parameters were set to [2, 1] and [0.0001, 0.0001].

\subsection{Experimental Results}

\subsubsection{Transferability of GE-AdvGAN++}

\begin{table*}[t]
\caption{Adversarial Success Rate of Generative Architecture Category Attack Methods}
\label{tab.geadvgan++}
\resizebox{\textwidth}{!}{%
\begin{tabular}{@{}c|c|ccccccc|c@{}}
\toprule
Model                       & Attack      & Inc-v3 & Inc-v4 & Inc-Res-v2 & Res-152 & \begin{tabular}[c]{@{}c@{}}Inc-v3\\ -adv-ens3\end{tabular} & \begin{tabular}[c]{@{}c@{}}Inc-v3\\ -adv-ens4\end{tabular} & \begin{tabular}[c]{@{}c@{}}Inc-Res\\ -adv-ens\end{tabular} & Average \\ \midrule
\multirow{3}{*}{Inc-v3}     & AdvGAN      & -      & 36.5   & 15.8       & 48.6    & 48.8                                                       & 48.8                                                       & 26.9                                                       & 37.6    \\
                            & GE-AdvGAN   & -      & 90.9   & 75.1       & 88.1    & 82.4                                                       & 79.7                                                       & 69.9                                                       & 81.0    \\
                            & GE-AdvGAN++ & -      & 91.7   & 77.4       & 81.3    & 85.6                                                       & 82.3                                                       & 75.7                                                       & 82.3    \\ \midrule
\multirow{3}{*}{Inc-v4}     & AdvGAN      & 65.2   & -      & 9.1        & 64.9    & 20.9                                                       & 54.3                                                       & 24.1                                                       & 39.8    \\
                            & GE-AdvGAN   & 88.4   & -      & 69.1       & 81.4    & 81                                                         & 74.1                                                       & 68.6                                                       & 77.1    \\
                            & GE-AdvGAN++ & 84.7   & -      & 68.6       & 74.7    & 84                                                         & 82.1                                                       & 74                                                         & 78.0    \\ \midrule
\multirow{3}{*}{Inc-Res-v2} & AdvGAN      & 37.8   & 33.9   & -          & 28.2    & 11.7                                                       & 14.1                                                       & 12.1                                                       & 23.0    \\
                            & GE-AdvGAN   & 87.4   & 83.4   & -          & 80.3    & 72.1                                                       & 63.3                                                       & 57.1                                                       & 73.9    \\
                            & GE-AdvGAN++ & 81.3   & 81.6   & -          & 72.9    & 84.3                                                       & 82.8                                                       & 74.6                                                       & 79.6    \\ \midrule
\multirow{3}{*}{Res-152}    & AdvGAN      & 44.2   & 38     & 21.1       & -       & 16.1                                                       & 24.9                                                       & 12.4                                                       & 26.1    \\
                            & GE-AdvGAN   & 85     & 83.5   & 68.7       & -       & 77.9                                                       & 76.2                                                       & 67.4                                                       & 76.5    \\
                            & GE-AdvGAN++ & 78.5   & 80.1   & 65.2       & -       & 83.2                                                       & 84                                                         & 75.2                                                       & 77.7    \\ \midrule
\multirow{3}{*}{MaxViT-T}   & AdvGAN      & 11.3   & 9.8    & 1.2        & 8.5     & 9.2                                                        & 10.7                                                       & 4.8                                                        & 7.9     \\
                            & GE-AdvGAN   & 57.4   & 52.3   & 45.7       & 57.6    & 47                                                         & 45.4                                                       & 35.3                                                       & 48.7    \\
                            & GE-AdvGAN++ & 71.3   & 73.8   & 53.9       & 65.3    & 76.2                                                       & 73.7                                                       & 68.5                                                       & 69.0    \\ \bottomrule
\end{tabular}%
}
\end{table*}

In this section, we first analyze the ASR performance of GE-AdvGAN++ compared to other similar methods. As shown in Table~\ref{tab.geadvgan++}, GE-AdvGAN++ achieves an average improvement of 47.8 in ASR compared to our main competing algorithm AdvGAN. Additionally, compared to GE-AdvGAN, our optimized GE-AdvGAN++ also demonstrates enhanced performance, with an average improvement of 5.9.

\subsubsection{Compatibility of the GE-AdvGAN+ Framework}\label{exp.geframe}
\begin{figure*}[t]
    \centering
    \includegraphics[width=\linewidth]{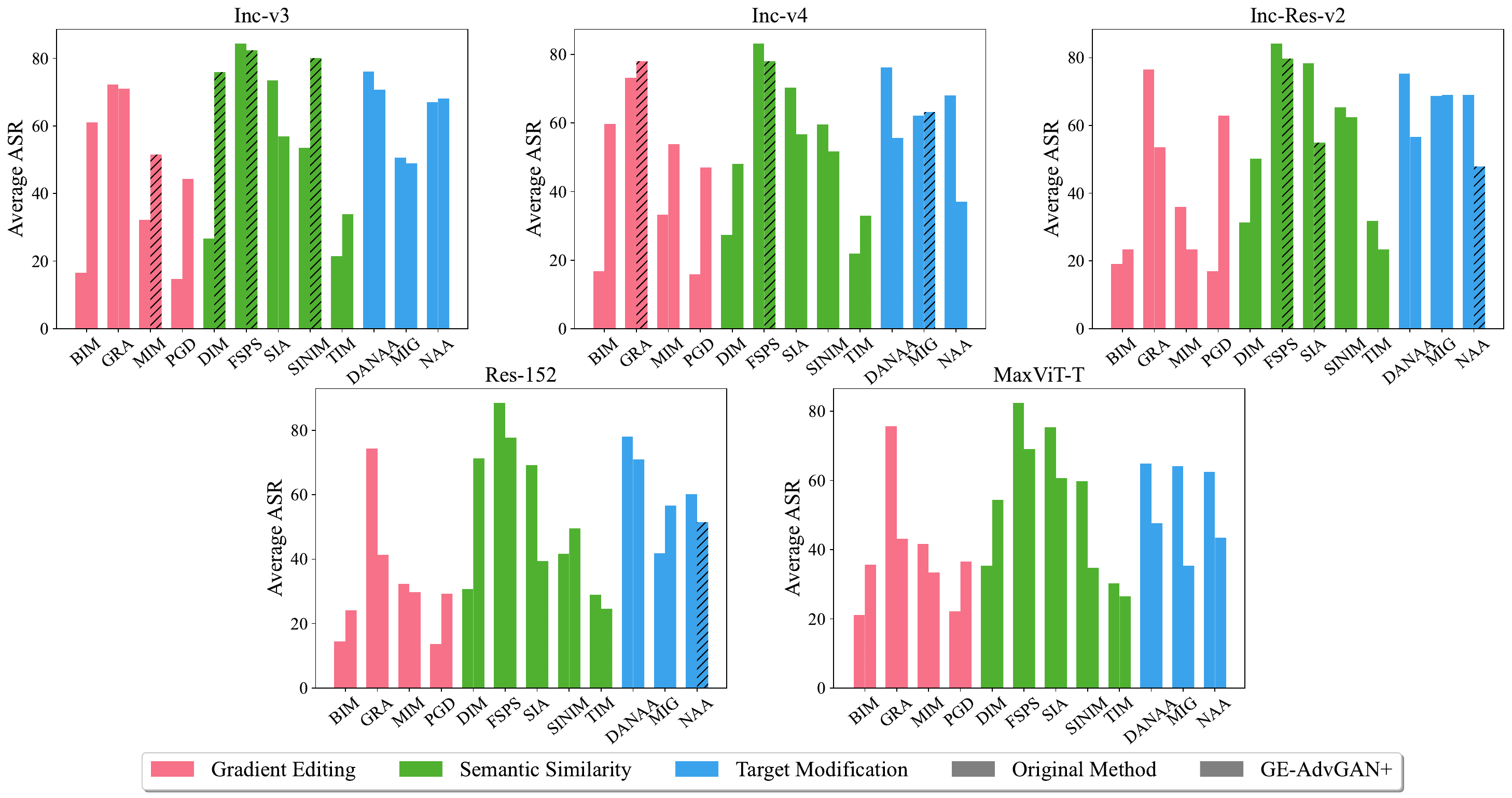}
    \caption{Comparison of the transferability of different adversarial attack methods with and without integration into the GE-AdvGAN+ framework. The five subfigures represent experiments with different models as surrogate models. The three colors represent three types of adversarial attack methods, with solid bars representing the original attack methods and shaded bars representing the performance of the GE-AdvGAN+ framework using the corresponding gradient information.}
    \label{fig.aasr}
\end{figure*}

In this section, we analyze the changes in transferability of the GE-AdvGAN framework when combined with almost all mainstream attack methods. As illustrated in Figure~\ref{fig.aasr}, we present the transferability of three types of adversarial attack methods and their performance when integrated with the GE-AdvGAN framework across five different surrogate models. The figure shows that the Gradient Editing methods significantly improve transferability when combined with GE-AdvGAN, particularly for BIM and PGD, which originally had relatively poor transferability. Moreover, the Semantic Similarity methods exhibit the best transferability performance when combined with GE-AdvGAN, with FSPS achieving the highest transferability across all models. However, when combined with Target Modification methods, traditional CNN models such as Inc-v3, Inc-v4, Inc-Res-v2, and Res-152 show a slight decrease in transferability, while the ViT model MaxViT-T exhibits a significant performance drop. These results highlight the potential enhancement effects of GE-AdvGAN across different adversarial attack methods and its adaptability to various model architectures.

\subsubsection{Computational Efficiency of GE-AdvGAN++}

In this section, we analyze the computational speed and resource consumption of GE-AdvGAN++ compared to other attack methods. In Table~\ref{tab.fps}, we compare the speed of generating adversarial samples, using Frames Per Second (FPS) as the evaluation metric. The results show that GE-AdvGAN++ significantly outperforms other methods in terms of speed, with an FPS of 2217.7, far exceeding other methods such as BIM (2.9 FPS) and MI-FGSM (2.83 FPS). This high efficiency is mainly attributed to GE-AdvGAN++'s ability to generate adversarial samples continuously after a single training, without requiring retraining for each attack.

\begin{table}[htpb]
\centering
\caption{Comparison of FPS rates across different adversarial attack generation methods.}
\label{tab.fps}
\resizebox{\textwidth}{!}{%
\begin{tabular}{@{}cccccccccccccc@{}}
\toprule
Method & BIM & DANAA & DI-FGSM & FSPS  & GRA   & MI-FGSM & MIG & NAA  & PGD   & SIA   & SINI-FGSM & TI-FGSM & GE-AdvGAN++ \\ \midrule
FPS    & 2.9 & 0.556 & 2.454   & 0.092 & 0.194 & 2.83    & 0.1 & 1.04 & 2.556 & 0.474 & 0.852     & 2.75    & \textbf{2217.7}      \\ \bottomrule
\end{tabular}%
}
\end{table}

Additionally, Table~\ref{tab.papams} presents a comparison of the model sizes in terms of parameters. The Generator used by GE-AdvGAN++ has a parameter size of only 0.33 MB, significantly smaller than other models such as Inc-Res-v2 (213.03 MB) and Res-152 (229.62 MB). This reduces the demand for storage resources and the usage of video memory when generating adversarial samples, thereby enhancing operational flexibility and efficiency, especially in edge computing applications.

\begin{table}[htpb]
\centering
\caption{Parameter sizes of different models utilized in adversarial attack generation.}
\label{tab.papams}
\resizebox{0.7\textwidth}{!}{%
\begin{tabular}{@{}lccccc@{}}
\toprule
Model            & Inc-Res-v2 & Inc-v3 & MaxViT-T & Res-152 & Generator \\ \midrule
Parameter size (MB) & 213.03     & 90.92  & 117.84   & 229.62  & 0.33      \\ \bottomrule
\end{tabular}%
}
\end{table}

\section{Conclusion and discussions}
In this study, we introduced the GE-AdvGAN+ framework, an advanced framework for generating highly transferable and computationally efficient adversarial attacks that are easy to deploy. By incorporating nearly all mainstream attack methods, GE-AdvGAN+ significantly enhances the transferability of adversarial attacks and the efficiency of sample generation. Our comprehensive evaluation across multiple datasets and models demonstrates that this framework offers superior transferability and computational efficiency compared to existing methods. Furthermore, our results highlight the framework's adaptability to different model types, including traditional CNNs and more complex architectures like ViTs. The reduction in parameter size and increase in FPS for GE-AdvGAN++ make it a viable option for practical applications, particularly in edge computing environments. Overall, GE-AdvGAN++ represents a significant advancement in the field of adversarial attacks, providing a comprehensive and efficient tool for researchers and practitioners. Future work will explore the integration of more advanced interpretability methods and the application of GE-AdvGAN++ in domains beyond image classification.

\bibliographystyle{ACM-Reference-Format}
\bibliography{main}










\end{document}